\documentclass{article}

\PassOptionsToPackage{compress}{natbib}

\usepackage[final]{nips_2018}

\usepackage[utf8]{inputenc} 
\usepackage[T1]{fontenc}    
\usepackage{hyperref}       
\usepackage{url}            
\usepackage{booktabs}       
\usepackage{amsfonts}       
\usepackage{nicefrac}       
\usepackage{microtype}      
\usepackage{soul}
\usepackage{xspace}
\usepackage{booktabs}
\usepackage{todonotes}
\usepackage{enumitem}
\usepackage{xcolor}
\usepackage{comment}
\usepackage{graphicx}
\usepackage{multirow}
\usepackage{amsmath,amsthm,amssymb}

\newcommand\TODO[1]{\textcolor{red}{\\TODO: #1\\}}

\newcommand{\twacg}{\texttt{\textsc{TextWorld ACG}}\xspace}

\newcommand{\GRU}{{\mathrm{GRU}}}

\newcommand{\acg}{\texttt{\textsc{ACG}}\xspace}
\newcommand{\acge}{\texttt{\textsc{ACGE}}\xspace}
\newcommand{\code}[1]{\texttt{#1}}
\newcommand{\cmd}[1]{\textbf{\small{\code{#1}}}}

\title{Towards Solving Text-based Games by Producing Adaptive Action Spaces}

\author{
    Ruo Yu Tao \\
    McGill University\\
    Microsoft Research \\
    \texttt{ruo.tao@mail.mcgill.ca} \\
    \And
    Marc-Alexandre Côté \\
    Microsoft Research \\
    \texttt{macote@microsoft.com} \
    \And
    Xingdi Yuan \\
    Microsoft Research \\
    \texttt{eric.yuan@microsoft.com} \\
    \And
    Layla El Asri \\
    Microsoft Research \\
    \texttt{layla.elasri@microsoft.com} \
}

\begin{document}

\maketitle

\begin{abstract}
To solve a text-based game, an agent needs to formulate valid text commands for a given context and find the ones that lead to success. Recent attempts at solving text-based games with deep reinforcement learning have focused on the latter, i.e., learning to act optimally when valid actions are known in advance. In this work, we propose to tackle the first task and train a model that generates the set of all valid commands for a given context. We try three generative models on a dataset generated with Textworld \citep{cote2018textworld}. The best model can generate valid commands which were unseen at training and achieve high $F_1$ score on the test set.

\end{abstract}

\section{Introduction}
Text-based games offer a unique framework to train decision-making models insofar as these models have to understand complex text instructions and interact via natural language. At each time step of a text-based game, the current environment of the player (or the `context') is described in words. To move the game forward, a text command (or an `action') must be issued. Based on the new action and the current game state, the game transitions to a new state and the new context resulting from the action is described to the player. This iterative process can be naturally divided into two tasks. The first task is to recognize the commands that are possible in a given context (e.g., \textit{open the door} if the context contains an unlocked door), and the second task is the reinforcement learning task of learning to act optimally in order to solve the game \citep{narasimhan2015lstmdqn, zelinka2018, he2015drrn, haroush2018actionelimination}. Most work on reinforcement learning has focused on training an agent that picks the best command from a given set of valid commands, i.e., pick the command that would lead to completing the game. 


Humans who play a text-based game typically do not have access to a list of commands and a large part of playing the game consists of learning how to formulate valid commands. In this paper, we propose models that try to accomplish this task. We frame it as a supervised learning problem and train a model by giving it (input, label) pairs where the input is the current context as well as the objects that the player possesses, and the output is the list of \emph{admissible commands} given this input. Similarly to \citet{cote2018textworld}, we define an admissible command as a command that changes the game's state. We generate these (input, label) pairs with TextWorld ~\citep{cote2018textworld}, a sandbox environment for generating text-based games of varying difficulty.

In this work, we explore and present three neural encoder-decoder approaches:
\begin{itemize}[noitemsep,nolistsep]
\item a pointer-softmax model that uses beam search to generate multiple commands;
\item a hierarchical recurrent model with pointer-softmax generating multiple commands at once;
\item a pointer-softmax model generating multiple commands at once.
\end{itemize}

The first model has the disadvantage of imposing a fixed number of actions for any given context. The two others alleviate this constraint but suffer from conditioning on the previous action generated. We compare empirical and qualitative results from those models, and pinpoint their weaknesses.
\begin{figure}
  \centering
  \includegraphics[scale=0.28]{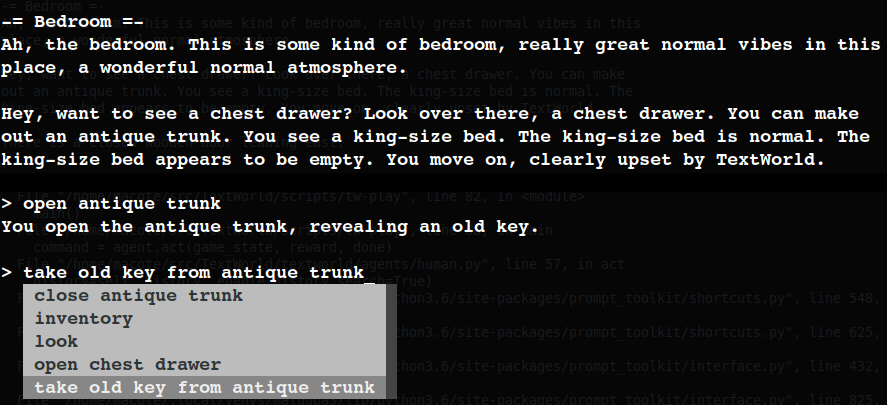}
  \caption{An instance of a TextWorld game. The context is the concatenation of the room's and inventory's description, and the admissible commands are the actions that would affect the state.}
  \label{fig:example_game}
\end{figure}

\section{Related Work}

\begin{table}[!t]
    \centering
    \scriptsize
    \begin{tabular}{r|r|r|r|l}
        \toprule
        Dataset & Train & Valid & Test & Target\\
        \midrule
        \text{\scriptsize \acg} & 33,716 & 4,243 & 4,276 & 9.37 $\pm$ 26.3 \\
        \text{\scriptsize \acge} & 214,741 & 26,904 & 27,423 & 2.47 $\pm$ 0.7 \\
        \bottomrule
    \end{tabular}
    \quad
    \begin{tabular}{r|r|r|r}
        \toprule
        Dataset & Commands & Unique & Unseen\\
        \midrule
        \text{\scriptsize Train} & 315,974 & 14,328 & - \\
        \text{\scriptsize Valid} & 39,302 & 4,006 & 846 \\
        \text{\scriptsize Test}  & 40,464 & 4,243 & 921 \\
        \bottomrule
    \end{tabular}
    \caption{Statistics of the datasets. Left) Number of data points in each train/valid/test split, and the average number of admissible commands (i.e., target) per data point. Right) Number of commands in each train/valid/test split, how many of those are unique, and how many are not seen in the train set.}
    \label{tab:twacg_stats}
\end{table}
\paragraph{Sequence to sequence generation}
Sentence generation has been studied extensively with the inception of sequence to sequence models \citep{sutskever2014seq2seq}, and attentive decoding \citep{bahadanau2014attention}. Pointer-based sequence to sequence networks \citep{vinyals2015pointer, gulcehre2016pointersoftmax, wang2017qa} help dealing with out-of-vocabulary words by introducing a mechanism for choosing between outputting a word from the vocabulary or referencing an input word during decoding. \citet{vinals2015order} studied the problem of matching input sequences to output sets, i.e., where there is no natural order between the elements. This task is challenging because there is no natural order between the sentences but there is an order between the tokens within each sentence. One of the models we try for this task is the hierarchical encoder-decoder \citep{sordoni2015hred} originally proposed to model the dialogue between two speakers. Another model is inspired by \citet{yuan2018keyphrase} who generates concatenated target sentences with orthogonally regularized separators.

\vspace{-0.5em}
\paragraph{Reinforcement learning for text-based games}
Many recent attempts at solving text-based games have assumed that the agent has a predefined set of commands to choose from. For instance, the Action-Eliminating Network ~\citep{haroush2018actionelimination} assumes that the agent has access to all possible permutations of commands in the entire game, and prunes that list in each state to allow the agent to better select correct commands. One attempt at command generation for a text-based game is the LSTM-DQN ~\citep{narasimhan2015lstmdqn}. This approach generates commands by leveraging off-policy deep Q-value approximations ~\citep{mnih2013dqn}, and learns two separate Q-functions for verbs and nouns. This limits the structure of generated commands to verb-action pairs, and does not allow for more robust multi-entity commands. \cite{yuan2018twcount} extends the LSTM-DQN approach with an exploration bonus to try and generalize, and beat games consisting of collecting coins in a maze.

\vspace{-0.5em}
\paragraph{Separating planning from generation in dialogue systems}
The task of choosing the best next utterance to generate for a given context has been extensively studied in the literature on dialogue systems \citep{rieser_natural_2016,Pietquin:11,Fatemi:16}. Historically, dialogue systems have considered separately the tasks of understanding the context, producing the available next utterances and of generating the next utterance \citep{Lemon:07}. Recent attempts at learning to perform all these tasks through one end-to-end model have produced encouraging results \citep{li_adversarial_2017,Bordes:17} but so far, the best-performing models still separate these two tasks \citep{Wen:16,Asadi:16}. Inspired by these results, we decide to frame the task of solving a text-based game into an action generation and an action selection modules and we propose models for action generation in the following section.

\section{Methodology}

\subsection{Dataset and environment}

In this section, we introduce a dataset called TextWorld Action Command Generation (\twacg). It is a collection of game walkthroughs gathered from random games generated with TextWorld. 
Statistics of \twacg are shown in Table~\ref{tab:twacg_stats}.
Each data point in \twacg consists of:
\begin{enumerate}[noitemsep,nolistsep]
    \item Context: concatenation of the room's and inventory's description for a game state;
    \item Entities: a list of interactable object names or exits appearing within the context;
    \item Commands: a list of strings that contains all the admissible commands recognized by TextWorld.
\end{enumerate}

We define two tasks using \twacg to learn the action space of these TextWorld games. First, without conditioning on entities, the model needs to generate all the admissible commands. Second, conditioning on one entity, the model is required to generate all valid commands that are related to that entity. In the following sections, we denote the task without conditioning on entity with \acg, and the task conditioning on entities with \acge. The data used for the \acge task is created by splitting each data point in \twacg by its entities, so that each data point in \acge has a single entity. There exist commands with multiple entities (i.e., \cmd{put apple on table}) - in these cases we group this action with one of the entities, and expect the models to produce the other entity. We also ignore the two commands (\cmd{look} and \cmd{inventory}) that don't affect the game state. This is because the context already consists of the exact descriptions returned by \cmd{look} and \cmd{inventory}. Adding the two commands would only serve to inflate metrics.

\subsection{Command generation}

In the following sections, we denote tokenized input words from the context sequence as $w$, $x$ to denote embedded tokens, a subscript ($e$ or $d$ etc.) to denote where the representations are from (encoder, decoder etc.), $h$ to represent hidden states, $s$ to represent session states and $y$ to denote output tokens. We use superscripts to represent time steps. An absence of a superscript represents multiple time-steps. We represent concatenation with angled brackets $\langle \ \rangle$. We also represent linear transformations with $L$, as well as linear transformations followed by an non-linear activation function $f$ as $L^f$. A subscript on these transformations (ie. $L_1, L_2$) represent transformations with different parameters.

\subsubsection{Context encoding}
Given a sequence of length $N$ in the context, we have the input sequence $w = (w_e^1, \dots, w_e^N)$ which we embed using GloVe ~\citep{pennington2014glove} vectors to produce $x_e = (x_e^1, \dots, x_e^N)$. We feed $x_e$ into a bidirectional RNN ~\citep{cho2014nmt, schuster1997brnn} to retrieve forwards ($h_{e, f}$) and backwards ($h_{e, b}$) encodings of the source sequence:
\begin{equation}
\begin{aligned}
\small
h^t_{e, f} &= \GRU_{e, f}(x_e^t, h_{e, f}^{t - 1}),\\
h^t_{e, b} &= \GRU_{e, b}(x_e^t, h_{e, b}^{t + 1}).
\end{aligned}
\end{equation}
We concatenate the two to get the resulting encoded sequence $h^t_e = \langle h^t_{e, f}, h^t_{e, b} \rangle$.
Then, we take a step depending on whether we condition on entity or not. Given a sequence of $m$ word tokens $(w_{ent}^1, \dots, w_{ent}^m)$ from the entity (which is also a sequence of word tokens), we find the indices $0 < i < j < N$ where the entity words appear in context, i.e., $(w_{ent}^1, \dots, w_{ent}^m) = (w_e^i, \dots, w_e^j)$. Now we take context encodings $(h^i_e, \dots, h^j_e)$ and use them as input to a GRU, where $i \leqslant t \leqslant j$:
\begin{equation}
\small
    h^t_{ent} = \GRU_{ent}(h^t_e, h^{t - 1}_{ent}).
\end{equation}
We use the final hidden state of this entity RNN as an entity encoding, which we will label as $h_{ent}$.

\begin{figure}
  \centering
  \includegraphics[scale=0.35]{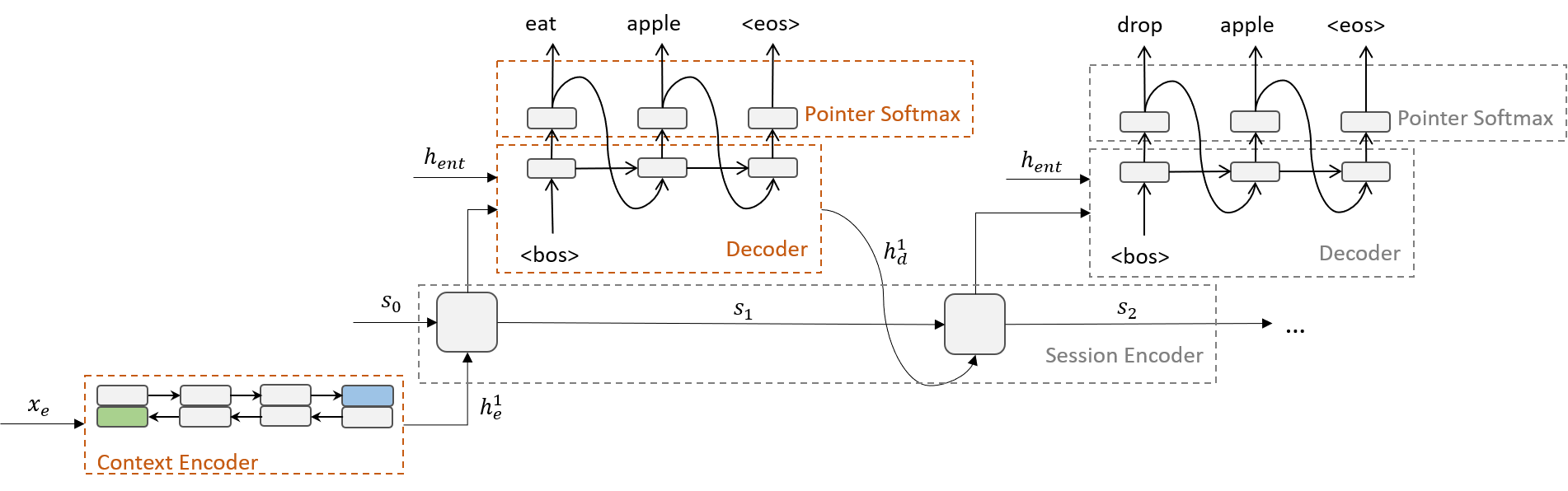}
  \caption{Hierarchical decoding of multiple commands given a single context. The orange outline represents the portions of the model also used for PS + BS and PS + Cat architectures.}
  \label{fig:model_architecture}
\end{figure}

\subsubsection{Attentive decoding and Pointer Softmax}
The decoder is also a recurrent model that takes in the context encodings $h_e$ and the generated entity encoding $h_{ent}$, at every timestep $t$ it produces a probability distribution of generating the next token $p(y_t)$. This next token can come from one of two sources - either a word in the context or a word in our shortlist vocabulary. Our shortlist vocabulary in this case is just our entire vocabulary (consisting of all possible 887 unique words in the dataset). The first part of the decoder model is an RNN that takes in the embedding of the previous output $x_d^{t} = embed(y^{t - 1})$ and previous decoder hidden state $h_d^{t - 1}$ to produce the first hidden state:
\begin{equation}
\small
    h_{d1}^t = \GRU_{d1}(x_d^{t}, h_d^{t - 1})
\end{equation}
Next, we concatenate this output hidden state with the entity representation to produce $u^t = \langle h_{d1}^t, h_{ent}\rangle$. We use this as the query to an attention mechanism~\citep{bahadanau2014attention} which generates annotations from this query and a value (in this case context encodings $h_e$). We generate these annotations with a two layer Feed Foward Network (FFN), and define a distribution over the context encodings. The context vector $c^t$ is then computed by taking the weighted sum of the context encodings $h_e$:
\begin{equation}
    \begin{aligned}
    \small
        \alpha^t &= softmax(L_1(L_2^{tanh}(\langle u^t, h_e \rangle)))\\
        c^t &= \sum_{i}\alpha^{t, i}h_e^i
    \end{aligned}
\end{equation}
We now use the annotations as the distribution over the context sequence, $p_c(y^t) = \alpha^t$. We take the context vector and use this and the first RNN hidden state $h_{d1}^t$ as input to a second RNN:
\begin{equation}
\small
    h_{d2}^t = \GRU_{d2}(c^t, h_{d1}^{t})
\end{equation}
We use this hidden state as the previous hidden state for the next time step ($h_d^t = h_{d2}^t$). We also apply dropout on the output of this RNN for regularization purposes. We now use the concatenation of $\langle h_{d2}^t, c^t, x_d^{t} \rangle$ as input to both the shortlist FFN and switch FFN to generate the shortlist distribution $p_s(y^t)$ and switch distributions $s^t$ respectively:
\begin{equation}
    \begin{aligned}
    \small
    p_s(y^t) &= softmax(L_3(L_4^{tanh}(\langle h_{d2}^t, c^t, x_d^{t} \rangle)))\\
    s^t &= sigmoid(L_5(L_6^{tanh}(\langle h_{d2}^t, c^t, x_d^{t} \rangle)))
    \end{aligned}
\end{equation}
We generate output tokens from a combined distribution over the context words ($p_c$), shortlist words and a switch that interpolate the probability of each distribution as per the Pointer Softmax ~\citep{gulcehre2016pointersoftmax} decoder framework
\begin{equation}
\small
    p(y^t) = s^t\cdot p_s(y^t) + (1 - s^t)\cdot p_c(y^t).
\end{equation}

\subsubsection{Hierarchical session encoding}
We adopt the framework of the hierarchical recurrent encoder-decoder ~\citep{sordoni2015hred} as one solution to alleviate the problem of multiple phrase generation per context (Figure~\ref{fig:model_architecture}). We place the session-level RNNs in between the encoder and decoder in order to condition on and summarize the previously decoded phrases. The session-level RNN takes as input a sequence of query representations $q^1, \dots, q^M$. We let $q^1 = h^N_e$, and all subsequent $q^i$'s will be the final decoder hidden state as per Figure~\ref{fig:model_architecture}. The session-level state becomes $session^m =\GRU_{ses}(session^{m - 1}, q^m)$, which we use as initial hidden states of the decoder, $h_d^0 = session^m$.

\subsubsection{Learning with command generation}
We employ a cross-entropy loss for all the learning objectives. The first model architecture uses the context encoder connected with a pointer-softmax decoder on single target commands (we label this as PS + BS($k$, $W$). During inference, we use the top $k$ out of $W$ beams to produce $k$ commands. With $S^i$ as the phrase produced at time step $i$, we try to maximize the following log-likelihood:
\begin{equation}
\small
    \mathcal{L}(S^i) = \sum_{t = 1}^{T}\log p(y^{i, t} \ | \ y^{i, 1:t-1}),\ \text{for} \ i = 1, \dots, |S|
\end{equation}
The second model applies hierarchical decoding, where we encode our context sequence as above and have a session state run through all the pointer-softmax decoder steps (we label this as HRED + PS). We use the same objective function as~\cite{sordoni2015hred} over the parameters for all the RNNs in the model. We let $S$ be all generated phrases given a context, $Q^m$ represents the phrase generated at session time-step $m$, so objective is to maximize the following log-likelihood:
\begin{equation}
\small
    \mathcal{L}(S) = \sum_{m = 1}^{M}\sum_{t = 1}^{T}\log p(y^t \ | \ y^{1:t-1}, Q^{1:m - 1})
\end{equation}
The final model uses the same architecture as \cite{yuan2018keyphrase}, we train on the concatenated target commands delineated by separator tokens (we label this as PS + Cat). In this case, the objective is:
\begin{equation}
\small
    \mathcal{L}(S) = \sum_{t = 1}^{|S|}\log p(y^t \ | \ y^{1:t-1})
\end{equation}

\section{Results and Discussion}
\begin{table}
    \centering
    \scriptsize
    \begin{tabular}{c|l|c|c|c|c|c}
        \toprule
        Dataset & Model & Precision & Recall & $F_1$ score & Unseen recall & Seen recall\\
        \midrule
        \multirow{3}{*}{\acg} &
        \text{PS + BS(11, 30)} & 26.6 & 53.3 & 35.5 & 12.0 & 54.8\\
        &\text{HRED + PS} & 94.1 & 84.7 & 89.2 & 48.4 & 86.0\\
        &\text{PS + Cat} & 98.4 & 94.7 & \textbf{96.5} & \textbf{83.0} & 95.1\\
        \midrule
        \multirow{3}{*}{\acge} &
        \text{PS + BS(3, 10)} & 20.1 & 93.0 & 33.0 & \textbf{80.9} & 93.5\\
        &\text{HRED + PS} & 96.8 & 91.7 & 94.2 & 59.7 & 92.8\\
        &\text{PS + Cat} & 98.9 & 96.3 & \textbf{97.6} & 76.7 & 96.9\\
        \bottomrule
    \end{tabular}
    \medskip
    \caption{Models performance on the \acg and \acge tasks. Where $k$ in PS + BS($k$, $W$) was determined by the highest $F_1$ score on the respective validation sets. The recall for commands seen (resp. unseen) during training is also reported.}
    \label{tab:twacg_results}
\end{table}
The empirical results (Table~\ref{tab:twacg_results}) and qualitative results (Appendix~\ref{appendix:qualitative}) show the ability for our best model to generate valid unseen commands and achieve $F_1$ scores of 96.5 and 97.6 on \acg and \acge respectively. The hierarchical and concatenation models outperform the Pointer-Softmax with Beam Search by a wide margin - largely due to the over-generation of PS + BS and the mismatch in number of targets between $k$ and actual number of target target commands (as seen in Appendix~\ref{appendix:dataset}). We hypothesize the PS + Cat outperforms the HRED model due to the gating mechanism between each session state. Conditioning on different queries gives HRED the ability to prevent gradients to flow through to the next session. We can see the detriment of this gating by comparing their $F_1$ scores. We hypothesize that as we only have a single query from our encoded context (and hence no "noisy" queries ~\citep{sordoni2015hred}) the gating mechanism hinders the model by "filtering" certain queries. We also observe a noticeable gap between the performances in the \acg and \acge as expected. In the \acge case, the models are more constrained by conditioning information. This means the scope of its generation narrows - our models generate smaller sequences on average (as shown in Table~\ref{tab:twacg_stats}), which decreases the likelihood of generating missing or extra commands as shown in Appendix~\ref{appendix:graphs}.

Experiments for models initialized without pre-trained GloVe embeddings were also conducted on both \acg and \acge datasets, but resulted in an almost negligible ($\leqslant 0.2\%$) decline in F1-score of the model. We postulate this is due to the mismatch in objectives between how GloVe is trained and the required entity relations in our environment. 

Interestingly, the generative models are able to generate a large portion of the valid commands that are unseen during training. Added diversity from beam search seems to help in producing unseen examples, but only in the case where the number of targets for a training instance is close to the number of targets we generate during inference as seen in Table~\ref{tab:twacg_results}. A large beam width is able to generate more unseen actions because of how beam search over generates actions.

In this work, we explored three different approaches at generating sets of text commands that are context dependent. We tested them on \twacg and \acge, two new datasets built using TextWorld. Seeing those encouraging results, our next step would be to combine the command generation with a control policy in order to play (and solve) text-based games. While the performance of the command generation is good (on TextWorld games), using it as a fixed generator would set an upper bound on the performance of the control policy (i.e., commands, mandatory for the game progression, might never be generated in the first place). Instead, our next goal is to develop a control policy that can use the generator and fine tune it to produce more relevant commands.


\section*{Acknowledgments}
Special thanks to Kaheer Suleman for his help and guidance in model architectures.


\medskip

\bibliography{biblio}
\bibliographystyle{apalike}

\clearpage
\appendix

\section{Full Results}

\subsection{Qualitative results from generation}
\label{appendix:qualitative}
\begin{table*}[h!]
    \centering
    \footnotesize
    \begin{tabular}{r|l}
        \toprule
        Context & -= attic = - you 've entered an attic .   you see a closed type p box . oh wow ! is that what i  \\
               & think it is ? it is ! it 's a workbench . you see a type p keycard and a bug on the workbench . \\
               & hmmm ... what else , what else ? there is an unblocked exit to the east . you do n't like \\
               & doors ?  why not try going south , that entranceway is unblocked .   you are carrying nothing . \\
        \midrule
        PS + BS  & \textbf{go bug}; go east; go south; \textbf{go type}; \textbf{open bug}; \textbf{open east}; \textbf{open type}; \textbf{open type p}; \\
        & open type p box; \textbf{open type p keycard'}; \textbf{take bug}; \textbf{take bug p keycard from}; \textbf{take east}; \\ 
        & \textbf{take south}; \textbf{take type}; \textbf{take type p}; \textbf{take type p box}; \textbf{take type p keycard}; \\
        & \textbf{take type p keycard from}\\
        \midrule
        HRED + PS  &  go east; go south; open type p box; \textit{take type p keycard from workbench};  \\
        \midrule
        PS + Cat  &  go east; go south; open type p box; \textit{take bug from workbench}; \\
        & \textit{take type p keycard from workbench};\\
        \midrule
 Ground Truth  &   go east; go south; open type p box; \textit{take bug from workbench}; \\
        & \textit{take type p keycard from workbench} \\
        \bottomrule
    \end{tabular}
    \caption{Example from \acg test set, predictions generated by 3 models. 
    All mismatched commands are shown in bold. All italicized commands are commands that are unseen during training.}
\end{table*}

\subsection{Full empirical results}
\label{tab:twacg__full_results}
\begin{table}[h!]
    \centering
    \scriptsize
    \begin{tabular}{r|c|c|c|c|c|c}
        \toprule
        & Dev & Test & Dev & Test & Dev & Test \\
        \midrule
        Model & \multicolumn{2}{c|}{\scriptsize Precision} & \multicolumn{2}{c|}{\scriptsize Recall} & \multicolumn{2}{c}{\scriptsize $F_1$ score} \\
        \midrule
        \text{PS + BS} & 26.5 & 26.6 & 54.1 & 53.3 & 35.6 & 35.5 \\
        \text{HRED + PS} & 94.6 & 94.1 & 85.6 & 84.7 & 89.9 & 89.2 \\
        \text{PS + Cat} & - & 98.4 & - & 94.7 & - & 96.5 \\
        \bottomrule
    \end{tabular}
    \quad
    \begin{tabular}{r|c|c|c|c|c|c}
        \toprule
        & Dev & Test & Dev & Test & Dev & Test \\
        \midrule
        Model & \multicolumn{2}{c|}{\scriptsize Precision} & \multicolumn{2}{c|}{\scriptsize Recall} & \multicolumn{2}{c}{\scriptsize $F_1$ score} \\
        \midrule
        \text{PS + BS} & 19.9 & 20.1 & 93.0 & 93.0 & 32.7 & 33.0 \\
        \text{HRED + PS} & 96.8 & 96.8 & 91.9 & 91.7 & 94.3 & 94.2 \\
        \text{PS + Cat} & - & 98.9 & - & 96.3 & - & 97.6 \\
        
        \bottomrule
    \end{tabular}
    \medskip
    \caption{Left: Model performance on the \acg task, with PS + BS we use the top $11$ beam search generated phrases. This number was determined by highest validation $F_1$ score. Right: Model performance on the \acge task. Again, PS + BS uses the top 2 commands generated by beam search, also determined by the highest validation $F_1$ score.}
\end{table}

\clearpage
\subsection{Graphical representation of results}
\label{appendix:graphs}
\begin{figure}[!h]
    \centering
    \includegraphics[width=.45\textwidth]{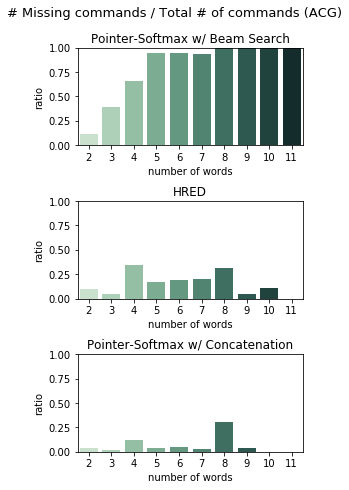}
    \includegraphics[width=.45\textwidth]{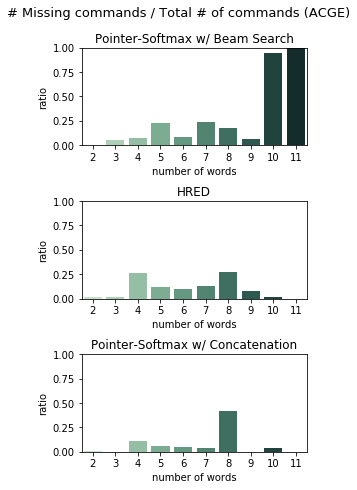}

    \caption{Ratio of missing commands by number of words in predictions to total commands by number of words in both \acg and \acge.}
\end{figure}
\quad
\begin{figure}[!h]
    \centering
    \includegraphics[width=.35\textwidth]{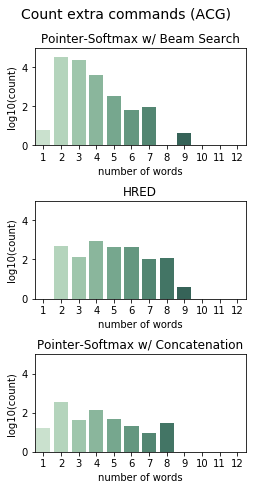}
    \includegraphics[width=.35\textwidth]{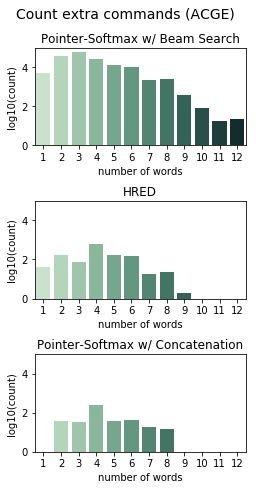}

    \caption{The count of extra commands generated by number of words in the command for both \acg and \acge.}
\end{figure}
\clearpage

\section{Dataset statistics}
\subsection{Additional statistics about the datasets}
\label{appendix:dataset}
\begin{figure}[!h]
    \centering
    \includegraphics[width=.95\textwidth]{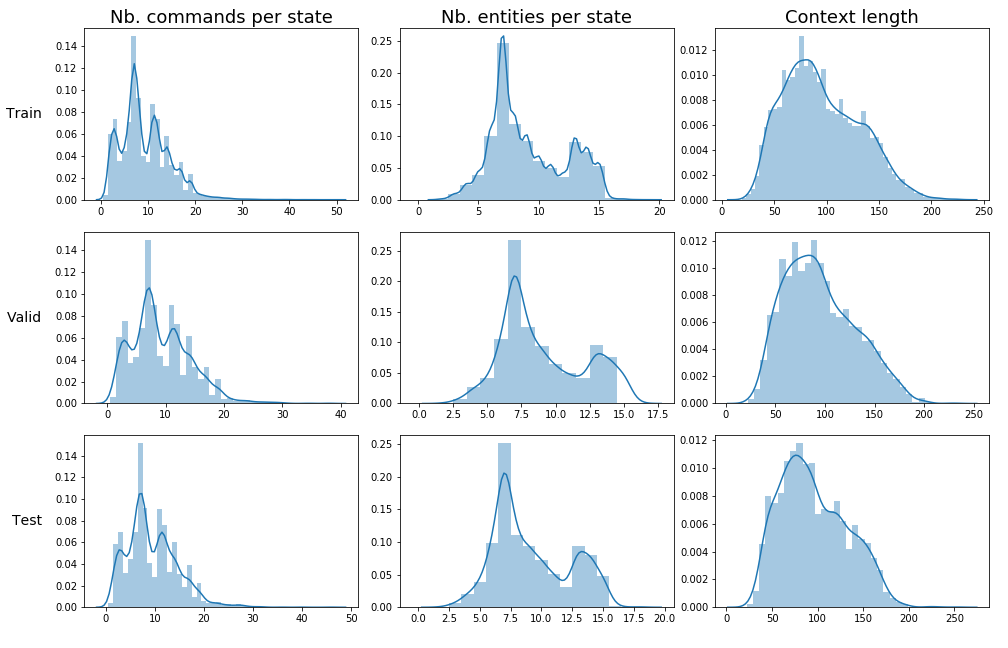}
    \caption{First column shows the number of admissible commands per extracted game state (i.e., represents the target in \twacg). Second column shows the number of entities (i.e., interactable objects or exits) per game state. Third column shows the number of words per context}
\end{figure}
\begin{figure}[!h]
    \centering
    \includegraphics[width=.95\textwidth]{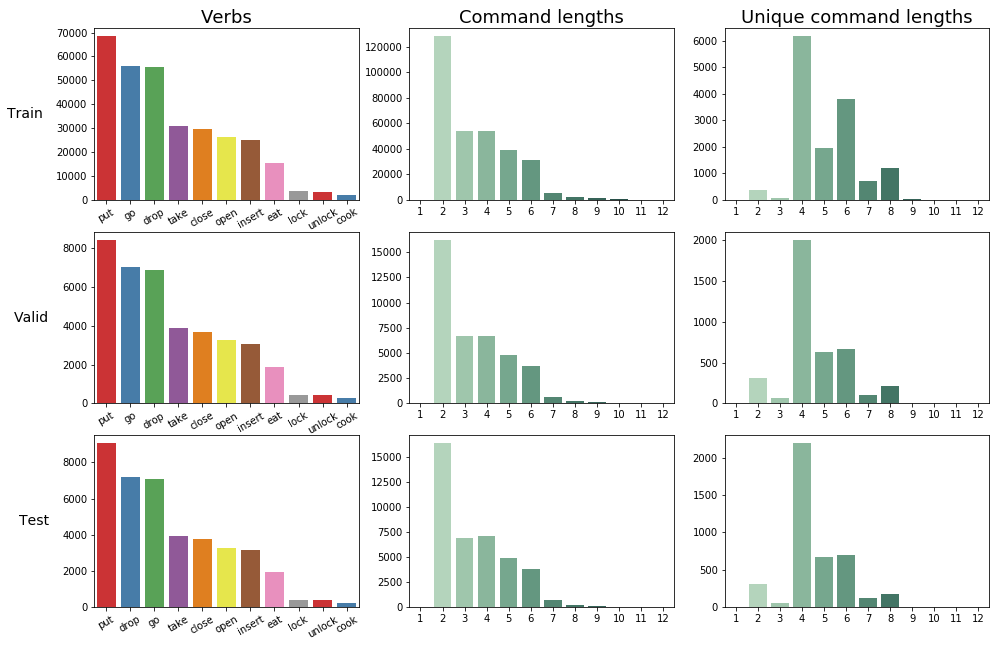}
    \caption{First column shows the frequencies of the verbs (i.e., first word of a command) in the dataset. Second column shows the length of the commands in the dataset. Third column shows the length of the unique commands.}
\end{figure}
\end{document}